\documentclass[10pt,twocolumn,letterpaper]{article}

\usepackage[pagenumbers]{cvpr}

\usepackage{makecell}
\usepackage{pifont}
\usepackage{cuted}

\definecolor{cvprblue}{rgb}{0.21,0.49,0.74}
\usepackage[pagebackref,breaklinks,colorlinks,allcolors=cvprblue]{hyperref}
\usepackage{array}
\usepackage{tabularx}
\usepackage{booktabs}
\usepackage{xcolor}
\usepackage{pifont}
\usepackage{makecell}
\usepackage{adjustbox}
\usepackage{graphicx}
\usepackage{multirow}
\usepackage{placeins}
\usepackage{amsmath}
\usepackage{listings}
\usepackage[most]{tcolorbox}
\usepackage{tikz}
\usepackage{fontawesome5}
\usepackage{varwidth}
\usepackage{pgfplots}
\pgfplotsset{compat=1.18}

\definecolor{chatgreen}{HTML}{95EC69}
\definecolor{chatgray}{HTML}{F2F2F2}
\definecolor{chatbg}{HTML}{F5F5F5}
\lstdefinestyle{promptstyle}{
    backgroundcolor=\color{gray!5},
    basicstyle=\ttfamily\footnotesize,
    breakatwhitespace=false,
    breaklines=true,
    captionpos=b,
    keepspaces=true,
    showspaces=false,
    showstringspaces=false,
    showtabs=false,
    tabsize=2,
    frame=single,
    rulecolor=\color{black!30}
}
\definecolor{darkgreen}{HTML}{006400}
\definecolor{lightgreen}{HTML}{00CC44}
\definecolor{darkblue}{HTML}{0A1F44}
\definecolor{darkred}{HTML}{8B0000}
\allowdisplaybreaks

\newcolumntype{Y}{>{\centering\arraybackslash}X}
\newcolumntype{R}{>{\raggedleft\arraybackslash}X}

\def\paperID{}
\def\confName{CVPR}
\def\confYear{2026}

\title{Learning to Wander: Improving the Global Image Geolocation Ability of LMMs via Actionable Reasoning}

\author{
    Yushuo Zheng\textsuperscript{\rm 1,2},
    Huiyu Duan\textsuperscript{\rm 1},
    Zicheng Zhang\textsuperscript{\rm 1,2},
    Xiaohong Liu\textsuperscript{\rm 1},
    Xiongkuo Min\textsuperscript{\rm 1}\\
    \textsuperscript{\rm 1}Shanghai Jiao Tong University \quad
    \textsuperscript{\rm 2}Shanghai Artificial Intelligence Laboratory\\
    {\tt\small \{yushuozheng, huiyuduan, zzc1998, xiaohongliu, minxiongkuo\}@sjtu.edu.cn}
}

\begin{document}
\maketitle
\setlength{\stripsep}{10pt plus 2pt minus 2pt}
\begin{strip}
    \vspace{-4em}
    \centering
    \includegraphics[width=\textwidth]{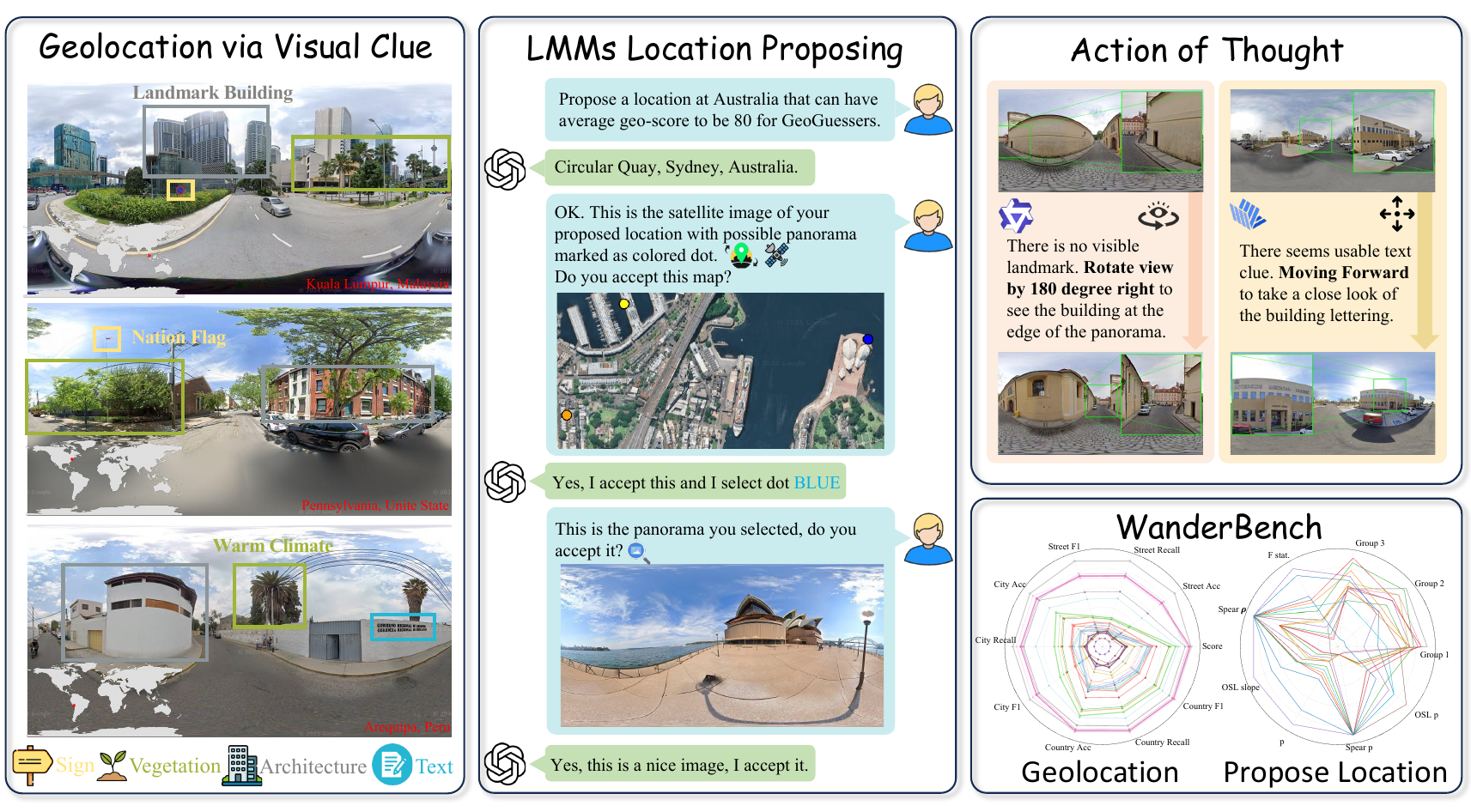}
    \captionsetup{hypcap=false}
    \captionof{figure}{Visual geolocation uses visual clues to determine the location of an image. WanderBench evaluates both the model’s ability to propose locations under different difficulty levels and its geolocation accuracy across diverse scenes. The Action of Thought allows the model to actively explore nearby views to gather more information for a more accurate result. GeoAoT further improves overall geolocation performance, shown by the thicker line in the radar chart along with location proposing performance. All radar chat is min max normalized.}
    \label{fig:heading}
\end{strip}
\begin{abstract}
Geolocation, the task of identifying the geographic location of an image, requires abundant world knowledge and complex reasoning abilities. Though advanced large multimodal models (LMMs) have shown superior aforementioned capabilities, their performance on the geolocation task remains unexplored. To this end, we introduce \textbf{WanderBench}, the first open access global geolocation benchmark designed for actionable geolocation reasoning in embodied scenarios. WanderBench contains over 32K panoramas across six continents, organized as navigable graphs that enable physical actions such as rotation and movement, transforming geolocation from static recognition into interactive exploration.
Building on this foundation, we propose \textbf{GeoAoT} (Action of Thought), a \underline{Geo}location framework with \underline{A}ction of \underline{T}hough, which couples reasoning with embodied actions. Instead of generating textual reasoning chains, GeoAoT produces actionable plans such as, approaching landmarks or adjusting viewpoints, to actively reduce uncertainty. We further establish an evaluation protocol that jointly measures geolocation accuracy and difficulty-aware geolocation questioning ability.
Experiments on 19 large multimodal models show that GeoAoT achieves superior fine-grained localization and stronger generalization in dynamic environments. WanderBench and GeoAoT define a new paradigm for actionable, reasoning driven geolocation in embodied visual understanding.
\end{abstract}
    
\FloatBarrier

\section{Introduction}

Geolocation, the task of determining the image’s geographical location of an image, is a cornerstone for applications anging including autonomous navigation, fact-checking to emergency response and cultural exploration \citep{Cheng2021, Chalvatzaras2023} \textit{etc}. The fundamental challenge of geolocation lies in its demand for advanced world-knowledge and reasoning capabilities. It requires interpreting a complex array of contextual clues such as architectural styles, road signage, flora, and other cultural markers to infer a precise location. This intricate process makes geolocation a significant benchmark for the reasoning capabilities of both artificial models and human experts \citep{Khan2024}. Though advanced abilities of LMMs, their performance on the geolocation remains unexplored.

Past efforts to automate geolocation have primarily focused on a static paradigm. Models like PlaNet \citep{Weyand2016a} and Im2GPS \citep{vo2017} are trained to classify static images into coarse geographical grid cells. Subsequent retrieval based methods have improved precision by matching query images against vast, geotagged databases \citep{Zhu2022, Muller-Budack2018}. However, this entire line of research overlooks a critical aspect of human problem-solving \textit{i.e.}, \textbf{active exploration}. When humans are lost, they do not simply stare at a single snapshot; they move, turn their heads to read a distant sign, or walk closer to a landmark. This ability to perform actions to gather new information is a key differentiator, and its absence in current models presents a significant performance bottleneck. This research gap is largely a consequence of limitations in existing datasets. As shown in Table~\ref{tab:dataset-comparison}, prominent datasets, from web-scraped collections like Im2GPS3K \citep{vo2017} and YFCC26K to massive map-service datasets like OSV-5M \citep{Astruc2024} and GeoComp, consist of discrete, static images. While valuable, they lack the environmental framework to support agent based actions. A model trained on this data can learn \textit{what} a location looks like, but not \textit{how} to explore it. This prevents the development and evaluation of truly interactive geolocation agents.

To bridge this fundamental gap, we introduce \textbf{WanderBench}, the first open access, global geolocation dataset designed for \textbf{embodied agent interaction}. As detailed in Table~\ref{tab:dataset-comparison}, our dataset provides a framework where an agent can execute a vocabulary of actions, including rotation and movement. This new benchmark enables a shift from static image-to-location mapping to a dynamic, exploratory process that more closely mimics human reasoning.

Furthermore, we propose GeoAoT (Action of Thought), a novel agent that integrates reasoning with action to leverage the LMMs to perform the geolocation. It can reason about visual information and then execute an action (\textit{e.g.}, ``Move forward 10 meters'') to acquire the missing data. Finally, to create a more robust benchmark, we introduce a novel evaluation protocol where the AI must not only solve geolocation tasks but also \textbf{assess and generate new problems}. The agent is tasked with providing locations and questions designed to test a specific difficulty level, demonstrating a meta-understanding of the geographical reasoning required.

Our contributions are three-fold:
\begin{itemize}
    \item We present \textbf{WanderBench}, a new, open access global geolocation dataset with 32K panoramas, uniquely built to support interactive agent actions like movement and rotation.
    \item We introduce \textbf{GeoAoT} a novel agent framework that moves beyond static Chain-of-Thought reasoning by generating and executing physical actions within the environment to actively gather information.
    \item We propose a new evaluation methodology where the agent’s capabilities are measured by its ability to both solve geolocation tasks and generate new, difficulty-calibrated questions.
\end{itemize}

\begin{table}[t]
\captionsetup{justification=raggedright, singlelinecheck=false} 
\caption{Comparison of geolocation datasets. \textbf{WanderBench} is the only openly available, globally scoped dataset providing both navigation support and bidirectional geolocation evaluation.}
\centering
\small 
\resizebox{\columnwidth}{!}{%
\begin{tabular}{l c c c c c c}
\toprule
\textbf{Dataset} & \textbf{Size} & \textbf{\makecell{Geographic \\ Coverage}} & \textbf{Source} & \textbf{\makecell{Open\\Access}} & \textbf{\makecell{Navigation\\Support}} & \textbf{\makecell{Bidirectional\\Evaluation}} \\ 
\midrule
Google-WS~\cite{clark2023} & 15k & Global & Map Service & \textcolor{red}{\ding{55}} & \textcolor{red}{\ding{55}} & \textcolor{red}{\ding{55}} \\
GMCP~\cite{zamir2014} & 105K & Local & Map Service & \textcolor{red}{\ding{55}} & \textcolor{red}{\ding{55}} & \textcolor{red}{\ding{55}} \\ 
StreetCLIP~\cite{haas2023} & 1M & Unknown & Map Service & \textcolor{red}{\ding{55}} & \textcolor{red}{\ding{55}} & \textcolor{red}{\ding{55}} \\
Im2GPS~\cite{hays2008} & 237 & Local & Web-Scraped & \textcolor{green}{\checkmark} & \textcolor{red}{\ding{55}} & \textcolor{red}{\ding{55}} \\
Im2GPS3K~\cite{vo2017} & 2997 & Local & Web-Scraped & \textcolor{red}{\ding{55}} & \textcolor{red}{\ding{55}} & \textcolor{red}{\ding{55}} \\
YFCC4K~cite{vo2017} & 4536 & Local & Web-Scraped & \textcolor{red}{\ding{55}} & \textcolor{red}{\ding{55}} & \textcolor{red}{\ding{55}} \\
YFCC26K~\cite{theiner2022} & 26k & Local & Web-Scraped & \textcolor{green}{\checkmark} & \textcolor{red}{\ding{55}} & \textcolor{red}{\ding{55}} \\
MP-16~\cite{larson2017} & 4.7M & Local & Web-Scraped & \textcolor{red}{\ding{55}} & \textcolor{red}{\ding{55}} & \textcolor{red}{\ding{55}} \\
OSV-5M~\cite{Astruc2024} & 5.1M & Global & Map Service & \textcolor{green}{\checkmark} & \textcolor{red}{\ding{55}} & \textcolor{red}{\ding{55}} \\
GeoComp~\cite{song2025geolocationrealhumangameplay} & 3.3M & Global & Map Service & \textcolor{green}{\checkmark} & \textcolor{red}{\ding{55}} & \textcolor{red}{\ding{55}} \\
\midrule
\textbf{WanderBench} & 31K & \textbf{Global} & Web-Scraped & \textcolor{green}{\checkmark} & \textcolor{green}{\checkmark} & \textcolor{green}{\checkmark} \\
\bottomrule
\end{tabular}%
}
\label{tab:dataset-comparison}
\end{table}
\begin{figure*}[t]
    \centering
    \includegraphics[width=0.9\textwidth]{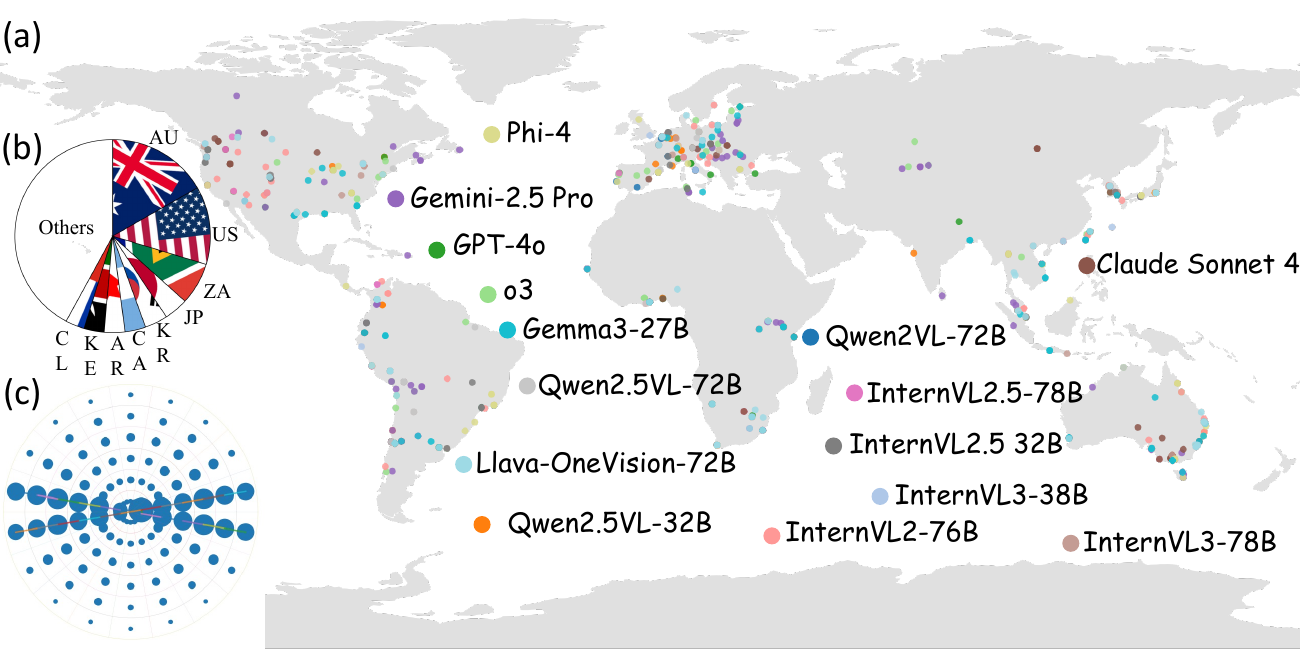}
    \caption{Figure 2.
(a) Global distribution of model-proposed locations across all continents.
(b) Country-level distribution of the proposed locations.
(c) Average navigation-graph structure of the WanderBench dataset.
}
    \label{fig:WanderBench}
\end{figure*}

\section{Related Work}
\subsection{Static Image Geolocation}

Image geolocation is the task of assigning geographic coordinates to an image has long challenged computer vision, spatial data mining, and GeoAI \citep{Zhu2023b, Zhu2023a, Zhu2024a, Zhao2017, Zhao2016, Zhang2023d, Han2023, Zhao2022, Zhang2023a, Zhang2023e}. Existing methods fall into two main paradigms: classification based and retrieval based. Classification models divide the Earth into grid cells and train neural networks to predict the cell of a given image \citep{clark2023, Pramanick2022, Muller-Budack2018, Seo2018, Weyand2016a}, enabling coarse localization but limiting fine-grained accuracy. Retrieval based methods instead match a query image to the most visually similar geotagged images in a reference database and infer coordinates from them \citep{Zhu2022, Muller-Budack2018, Zhang2022b, Workman2015, Liu2019, Zhou2024}, offering higher precision but requiring massive, densely populated databases and failing in sparse regions. Both paradigms, however, frame geolocation as a static, single-input problem, overlooking the dynamic, exploratory strategies humans use for spatial reasoning.

\subsection{Embodied Agents and Reasoning in Geolocation}

Research on embodied agents has primarily addressed goal directed navigation in simulated environments such as StreetLearn \citep{Mirowski2019}, where agents are trained to reach specific destinations rather than determine their own location. In contrast, Large Multimodal Models (LMMs) like GeoReasoner \citep{Li2024a} introduced the Geographical Chain of Thought, guiding models through step-by-step textual reasoning for static image based geolocation. While this improves interpretability, the reasoning remains passive: models can analyze what they see but cannot act to gather new evidence when information is missing or unclear. Our work bridges this divide by introducing an embodied agent that integrates reasoning with action, formulating and executing movement plans to actively collect disambiguating visual cues. Furthermore, unlike prior studies that only note variations in task difficulty \citep{Astruc2024}, our approach enables models to evaluate and generate tasks with controlled difficulty, advancing meta reasoning in geolocation.

\subsection{Geolocation Datasets}
Existing geolocation datasets inherently constrain model development due to issues of data quality, bias, and limited interactivity. Web-scraped datasets such as YFCC100M \citep{theiner2022} and Im2GPS3K \citep{vo2017} contain noisy, non-localizable content, while street view datasets like StreetLearn \citep{Mirowski2019} and others \citep{Astruc2024} offer higher visual quality but remain geographically biased or regionally restricted. More critically, all existing datasets are static comprising discrete images or pre rendered paths thus failing to support embodied or interactive geolocation tasks. To overcome this, we introduce \textbf{WanderBench}, a two-way benchmark that functions both as a dataset and as an evaluation framework. Beyond static accuracy, WanderBench assesses a model’s ability to propose diverse and valid locations under varying levels of difficulty. This dynamic, bidirectional evaluation design bridges perception and reasoning, enabling more comprehensive assessment of geospatial understanding in large multimodal models.

\vspace{-0.4em}
\section{WanderBench}
\subsection{Dataset Overview}

This dataset comprises 1,047 unique locations distributed across six continents, all generated by large multimodal models (LMMs). Each location is represented as a graph structure, where each node corresponds to a panoramic image with precise GPS coordinates, and edges represent navigable paths between these points. On average, each graph contains 31.27 panoramas (nodes) and 37.67 navigable connections (edges), resulting in a total of 32,741 panoramas and 39,442 edges across the entire dataset. Each panorama includes high resolution imagery and detailed metadata, supporting robust and fine grained geolocation research.

\subsection{Location Generation}
\label{sec:location_gen}
All benchmark locations are generated by large multimodal models (LMMs). We employ a two-stage process to ensure diversity and realism in the generated locations, as shown in Figure \ref{fig:heading}.
In the first stage, LMMs are prompted to produce detailed textual descriptions of regions within a specified continent and difficulty level, emphasizing distinctive geographical and cultural characteristics. Based on these prompts, the LMMs either output a textual description of a location or provide precise geographic coordinates.  
In the second stage, we present the LMMs with a satellite view centered on the described location at zoom level 16, where potential panorama positions are highlighted in distinct colors. If the LMMs reject the given map view, we regenerate the location description and repeat the process until a valid map view is obtained. Otherwise, the LMMs select one of the marked positions, and a corresponding panoramic image is retrieved for that site. If the selected panorama is deemed insufficient to represent the described location, the LMMs request additional panoramas from nearby marked positions. This iterative process continues until the LMMs determine that the set of panoramas accurately reflects the intended difficulty level and spatial characteristics.

\begin{table}[t]
\caption{Spatial diversity metrics of model-proposed locations. 
All coordinates normalized to $[0,1]^2$ per continent. 
$O_{16}$: occupancy on $16{\times}16$ grid; 
$H_{16}$: entropy; 
$A_{\text{hull}}$: convex-hull area; 
$R_{\text{CE}}$: Clark--Evans ratio; 
$\bar r_{\text{NN}}$: mean nearest-neighbor distance.}
\centering
\setlength{\tabcolsep}{4pt}
\resizebox{\columnwidth}{!}{%
\begin{tabular}{lccccc}
\hline
\multirow{2}{*}{\shortstack[c]{\textbf{Model}}}
& $O_{16}$ & $H_{16}$ & $A_{\text{hull}}$ & $R_{\text{CE}}$ & $\bar r_{\text{NN}}$ \\
& ($\times10^{-2}$) & ($\times10^{-1}$) & ($\times10^{-2}$) & ($\times10^{-1}$) & ($\times10^{-2}$) \\
\hline
InternVL2-76B~\cite{internvl2} & 2.73 & 3.10 & 8.61 & 3.21 & 4.63 \\
InternVL2.5-38B~\cite{internvl25} & 1.82 & 2.50 & 5.12 & 1.27 & 1.84 \\
InternVL2.5-78B~\cite{internvl25} & 1.95 & 2.58 & 4.31 & 1.63 & 2.35 \\
InternVL3-38B~\cite{internvl3} & 2.02 & 2.66 & 7.16 & 1.95 & 2.82 \\
InternVL3-78B~\cite{internvl3} & 2.08 & 2.62 & 5.40 & 1.97 & 2.84 \\
Phi-4~\cite{phi4} & 2.02 & 2.38 & 3.67 & 1.57 & 2.27 \\
Qwen2VL-72B~\cite{qwen2vl} & 1.95 & 2.57 & 3.22 & 0.95 & 1.38 \\
Qwen2.5VL-72B~\cite{qwen25vl} & 2.08 & 2.60 & 8.95 & 2.67 & 3.86 \\
Qwen2.5VL-32B~\cite{qwen25vl} & 2.02 & 2.62 & 4.05 & 1.42 & 2.04 \\
LLaVA-OV-72B~\cite{li2024llavaonevision} & 2.28 & 2.75 & 4.61 & 2.34 & 3.38 \\
Gemma-3-27B~\cite{gemma3} & 2.47 & 2.96 & 7.65 & 2.05 & 2.96 \\
Gemini-2.5 Pro~\cite{gemini25pro} & 2.73 & 3.23 & 5.70 & 2.42 & 3.50 \\
Claude Sonnet 4~\cite{claude4sonnet} & 1.89 & 2.27 & 3.20 & 1.61 & 2.33 \\
GPT-4o~\cite{gpt4o} & 2.41 & 2.92 & 6.23 & 2.75 & 3.96 \\
o3~\cite{o3} & 2.86 & 3.21 & 12.8 & 3.70 & 5.34 \\
\hline
\end{tabular}%
}
\label{tab:data-distribution}
\end{table}

\begin{figure}
    \centering
    \includegraphics[width=0.45\textwidth]{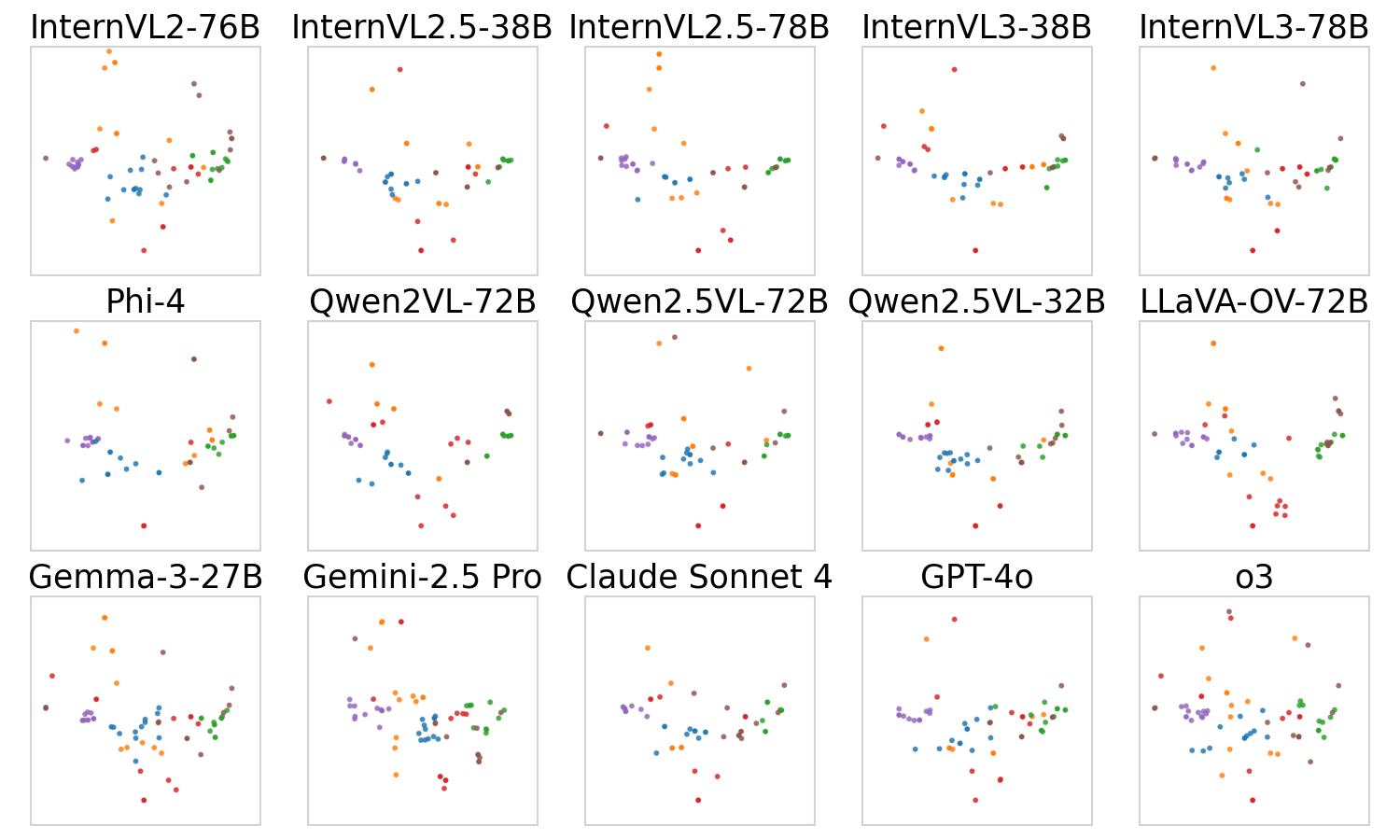}
    \caption{Visualization of the average navigation graph structure across all locations in the WanderBench dataset. Node sizes correspond to occurrence frequency within spatial bins, while edge thickness indicates transition frequency between bins.}
    \label{fig:PointDis}
\end{figure}
\subsection{Data Collection}
After the target locations are determined, we use map service provider APIs to retrieve panoramic imagery, GPS coordinates, and camera poses for each site. Each panorama is captured at a resolution of 2048$\times$1024 pixels, providing high quality visual data for geolocation and navigation tasks. The GPS coordinates are recorded with high precision to ensure accurate geospatial alignment.

For each region, we construct a navigation graph $G$ that represents the local street view topology:
\begin{align}
G &= (V, E), \label{eq:graph_def}\\
V &= \{v_1, v_2, \ldots, v_n\}, \label{eq:vertex_def}\\
E &= \{(v_i, v_j) \mid \text{a navigable path between } v_i \text{ and } v_j\}. \label{eq:edge_def}
\end{align}
Each node $v_i \in V$ corresponds to a panoramic image, while each edge $(v_i, v_j) \in E$ denotes a traversable connection between two panoramas.

To guarantee sufficient spatial continuity, we define the set of boundary nodes as
\begin{equation}
B = \{v \in V \mid \deg(v) = 1\}, \label{eq:boundary_def}
\end{equation}
and denote by $d(u, v)$ the shortest-path distance between any two nodes $u, v \in V$.  
Each graph is required to satisfy the following coverage constraint:
\begin{equation}
\min_{v \in V} \min_{b \in B} d(v, b) \geq 10, \label{eq:depth_constraint}
\end{equation}
ensuring that an agent starting from any node can explore at least ten steps without reaching a boundary or needing to backtrack.

This structured formulation produces spatially coherent graph representations that support robust navigation, exploration, and downstream geolocation analysis.
\begin{table*}[t]
\caption{Baseline performance of models on the benchmark. Lower is better for Distance (km), higher is better for all other metrics. \textcolor{blue}{\textbf{Blue bold}} values represent the best within the open-source model type.  \textcolor{red}{\textbf{Red bold}} values represent the best performance across all model types (closed-source).}
\centering
\setlength{\tabcolsep}{3pt}
\renewcommand{\arraystretch}{0.95}
\resizebox{\textwidth}{!}{%
\begin{tabular}{
    >{\centering\arraybackslash}m{1.9cm}
    >{\raggedright\arraybackslash}m{3.8cm}
    *{12}{>{\centering\arraybackslash}m{1.1cm}}
}
\toprule
\multirow{2}{*}{\textbf{Model Type}} &
\multirow{2}{*}{\centering\arraybackslash\textbf{Model}} &
\multicolumn{2}{c}{\textbf{Overall}} &
\multicolumn{3}{c}{\textbf{Street}} &
\multicolumn{3}{c}{\textbf{City}} &
\multicolumn{3}{c}{\textbf{Country}} \\
\cmidrule(lr){3-4}\cmidrule(lr){5-7}\cmidrule(lr){8-10}\cmidrule(lr){11-13}
& &
\textbf{Dist.\,$\downarrow$}  & \textbf{Score\,$\uparrow$} &
\textbf{Acc.\,$\uparrow$} & \textbf{Rec.\,$\uparrow$} & \textbf{F1\,$\uparrow$} &
\textbf{Acc.\,$\uparrow$} & \textbf{Rec.\,$\uparrow$} & \textbf{F1\,$\uparrow$} &
\textbf{Acc.\,$\uparrow$} & \textbf{Rec.\,$\uparrow$} & \textbf{F1\,$\uparrow$} \\
\midrule

\multirow{2}{*}{\textbf{\shortstack{Retrieval\\Based}}}
& GeoCLIP~\cite{geoclip}                        & 1925 & 65.81 & 0.158 & 0.158 & 0.158 & 0.302 & 0.302 & 0.302 & 0.611 & 0.611 & 0.611 \\
& GeoCLIP-top5                                 & 1403 & 73.55 & 0.232 & 0.233 & 0.233 & 0.392 & 0.392 & 0.392 & 0.700 & 0.700 & 0.700 \\
\midrule

\multirow{12}{*}{\textbf{\shortstack{Open\\Source\\LMMs}}}
& InternVL2.5-78B~\cite{internvl25}             & 1124 & 74.69 & 0.164 & 0.141 & 0.152 & 0.368 & 0.316 & 0.340 & 0.723 & 0.668 & 0.694 \\
& InternVL3-78B~\cite{internvl3}               & 1441 & 72.77 & 0.171 & 0.148 & 0.159 & 0.359 & 0.309 & 0.332 & 0.697 & 0.646 & 0.670 \\
& Qwen2VL-72B~\cite{qwen2vl}                   & 988.7  & 83.46 & 0.270 & 0.233 & 0.250 & 0.525 & 0.451 & 0.485 & 0.837 & 0.813 & 0.825 \\
& Qwen2VL-7B~\cite{qwen2vl}                    & 2531 & 69.02 & 0.110 & 0.094 & 0.101 & 0.341 & 0.292 & 0.315 & 0.675 & 0.625 & 0.649 \\
& Qwen2.5VL-32B~\cite{qwen25vl}                & 677.0  & 83.71 & 0.260 & 0.224 & 0.241 & 0.514 & 0.442 & 0.475 & 0.826 & 0.803 & 0.814 \\
& Qwen2.5VL-72B~\cite{qwen25vl}                & 674.8  & 84.58 & 0.257 & 0.221 & 0.238 & 0.550 & 0.473 & 0.509 & 0.833 & 0.810 & 0.821 \\
& Qwen2.5-VL-7B~\cite{qwen25vl}                & 1367 & 77.14 & 0.188 & 0.162 & 0.174 & 0.433 & 0.372 & 0.401 & 0.758 & 0.745 & 0.751 \\
& Gemma-3-27B~\cite{gemma3}                    & 1684 & 72.76 & 0.207 & 0.178 & 0.191 & 0.426 & 0.366 & 0.393 & 0.685 & 0.681 & 0.683 \\
& LLaVA-OV-72B~\cite{li2024llavaonevision}                  & 4080 & 53.68 & 0.112 & 0.097 & 0.104 & 0.241 & 0.206 & 0.222 & 0.508 & 0.482 & 0.494 \\
& LLaMA-4-Maverick~\cite{llama4maverick}       & 1731 & 72.73 & 0.198 & 0.199 & 0.198 & 0.372 & 0.372 & 0.372 & 0.696 & 0.695 & 0.696 \\
& LLaMA-3.2-90B-V~\cite{llama32_90b_vision}    & 1663 & 72.37 & 0.187 & 0.188 & 0.188 & 0.354 & 0.355 & 0.355 & 0.690 & 0.689 & 0.689 \\
& GLM-4.5V~\cite{glm45v}                       &
  \textcolor{blue}{\textbf{430.3}}  &
  \textcolor{blue}{\textbf{88.49}}   &
  \textcolor{blue}{\textbf{0.357}}   &
  \textcolor{blue}{\textbf{0.359}}   &
  \textcolor{blue}{\textbf{0.358}}   &
  \textcolor{blue}{\textbf{0.640}}   &
  \textcolor{blue}{\textbf{0.640}}   &
  \textcolor{blue}{\textbf{0.640}}   &
  \textcolor{blue}{\textbf{0.873}}   &
  \textcolor{blue}{\textbf{0.872}}   &
  \textcolor{blue}{\textbf{0.873}}   \\
\midrule

\multirow{7}{*}{\textbf{\shortstack{Closed\\Source\\LMMs}}}
& QwenVL-Max~\cite{qwenvlmax}                   & 1670 & 69.95 & 0.180 & 0.180 & 0.180 & 0.362 & 0.361 & 0.362 & 0.660 & 0.659 & 0.659 \\
& Claude Haiku 4.5~\cite{claude4sonnet}         & 2714 & 59.54 & 0.134 & 0.134 & 0.134 & 0.268 & 0.268 & 0.268 & 0.538 & 0.537 & 0.537 \\
& Claude Sonnet 4~\cite{claude4sonnet}          & 2136 & 66.01 & 0.138 & 0.139 & 0.139 & 0.270 & 0.271 & 0.271 & 0.594 & 0.593 & 0.593 \\
& Gemini-2.5 Flash~\cite{gemini25pro}           & 222.1  & 93.37 & 0.504 & 0.505 & 0.505 & 0.742 & 0.742 & 0.742 & 0.941 & 0.940 & 0.941 \\
& \textbf{Gemini-2.5 Pro~\cite{gemini25pro}}    &
  \textcolor{red}{\textbf{134.3}} &
  \textcolor{red}{\textbf{95.28}} &
  0.495 & 0.496 & 0.495 &
  \textcolor{red}{\textbf{0.783}} &
  \textcolor{red}{\textbf{0.784}} &
  \textcolor{red}{\textbf{0.783}} &
  \textcolor{red}{\textbf{0.958}} &
  \textcolor{red}{\textbf{0.958}} &
  \textcolor{red}{\textbf{0.958}} \\
& GPT-4o~\cite{gpt4o}                           & 313.7  & 91.90 & 0.403 & 0.404 & 0.404 & 0.730 & 0.730 & 0.730 & 0.919 & 0.919 & 0.919 \\
& \textbf{o3~\cite{o3}}                         & 318.8  & 92.01 &
  \textcolor{red}{\textbf{0.524}} &
  \textcolor{red}{\textbf{0.525}} &
  \textcolor{red}{\textbf{0.525}} &
  0.734 & 0.734 & 0.734 & 0.922 & 0.921 & 0.922 \\
\bottomrule
\end{tabular}%
}
\label{tab:baseline_results}
\end{table*}

\subsection{Questions Generation}
After the panoramas and navigation graphs are collected, we again leverage LMMs to generate geolocation questions for each location. For baseline task the LMMs are provided with the single panoramic images. They are prompted to create questions that require reasoning about spatial relationships, visual cues, and contextual information present in the panoramas. The generated questions vary in difficulty and type, including multiple choice, open ended, and scenario based queries. Each question is paired with a correct answer and, where applicable, a set of plausible distractors to facilitate comprehensive evaluation of geolocation models. This approach ensures that the questions are not only challenging but also relevant to the unique characteristics of each location, promoting advanced reasoning and exploration capabilities in geolocation tasks.

\subsection{Dataset Statistic}
Figure~\ref{fig:WanderBench}~(a) and (b) present the location distribution of our WanderBench dataset. The panoramas are collected from all six continents, with a balanced representation across different geographical regions from more than 50 countries. Each of LMM propose 4 location for each continent in three different difficulty level, in result each LMM will propose total 192 location. These panoramas span a diverse range of environments, from dense urban cityscapes to sparse rural landscapes, ensuring comprehensive coverage of various geographical and cultural features. The average edge to node ratio is 2.41, indicating well connected navigation graphs for realistic agent movement and exploration.

To better understand the global structural characteristics of the navigation graphs, we construct an average graph representation that aggregates spatial and topological patterns across all locations. Each graph is first normalized by translating its central node to the origin, scaling by its median edge length, and aligning its principal axis via PCA rotation. The normalized graphs are then projected into a unified polar coordinate space and discretized into radial and angular bins, where node occurrences and edge transitions are accumulated to derive a consensus topology. As shown in Figure~\ref{fig:WanderBench}(c), this visualization reveals dense central hubs, frequently traversed inter-bin connections, and approximately symmetric branching patterns. The size of each circle indicates the number of panoramas (nodes) within each spatial bin, and line thickness reflects the frequency of navigable transitions (edges), jointly capturing common spatial and topological structures across WanderBench. This consensus representation offers a compact yet expressive summary of the dataset’s overall navigability and typical spatial organization.

\begin{table*}[t]
\centering
\caption{ANOVA and trend analysis across coverage levels (20\%, 50\%, 80\%). Group 1--3 correspond to means at 20\%, 50\%, 80\%.}
\scriptsize
\setlength{\tabcolsep}{4pt}
\renewcommand{\arraystretch}{0.95}

\resizebox{\textwidth}{!}{%
\begin{tabular}{
    >{\centering\arraybackslash}m{1.7cm}
    >{\centering\arraybackslash}l
    *{9}{>{\centering\arraybackslash}m{1.1cm}}
    >{\centering\arraybackslash}m{1.1cm}
}
\toprule
\textbf{Model Type} &
\textbf{Model Name} &
\makecell{$F\,\uparrow$ \\ \textbf{Stat.}} &
\makecell{$p\,\downarrow$ \\ $(10^{-2})$} &
\makecell{$\rho\,\uparrow$ \\ \textbf{Spear}} &
\makecell{$p\,\downarrow$ \\ \textbf{Spear}} &
\makecell{\textbf{Slope}\,$\uparrow$ \\ \textbf{OLS}} &
\makecell{$p\,\downarrow$ \\ \textbf{OLS}} &
\makecell{\textbf{Avg.\ Score} \\ Hard} &
\makecell{\textbf{Avg.\ Score} \\ Medium} &
\makecell{\textbf{Avg.\ Score} \\ Easy} &
\textbf{Remark} \\
\midrule

\multirow{11}{*}{\textbf{Open Source}} 
 & InternVL2-76B~\cite{internvl2}         & 7.62 & 0.10 & 1.00 & 0.00 & 0.30 & 0.08 & 64.31 & 75.14 & 82.15 & \textcolor{green}{$\checkmark$} \\
 & InternVL2.5-38B~\cite{internvl25}      & 2.47 & 0.92 & 1.00 & 0.00 & 0.15 & 0.28 & 75.41 & 76.29 & 84.64 & \textcolor{green}{$\checkmark$} \\
 & InternVL2.5-78B~\cite{internvl25}      & 0.22 & 8.03 & 0.50 & 0.67 & 0.05 & 0.44 & 74.44 & 77.73 & 77.17 & \textcolor{red}{\ding{55}} \\
 & InternVL3-38B~\cite{internvl3}         & 2.81 & 0.67 & 1.00 & 0.00 & 0.15 & 0.24 & 78.52 & 80.03 & 87.80 & \textcolor{green}{$\checkmark$} \\
 & InternVL3-78B~\cite{internvl3}         & 0.55 & 5.82 & 1.00 & 0.00 & 0.07 & 0.27 & 75.64 & 79.60 & 80.07 & \textcolor{green}{$\checkmark$} \\
 & Phi-4~\cite{phi4}                       & 0.97 & 3.83 & 1.00 & 0.00 & 0.13 & 0.04 & 71.97 & 75.32 & 79.61 & \textcolor{green}{$\checkmark$} \\
 & Qwen2VL-72B~\cite{qwen2vl}             & 0.10 & 9.06 & 0.50 & 0.67 & 0.01 & 0.77 & 79.60 & 81.65 & 80.33 & \textcolor{red}{\ding{55}} \\
 & Qwen2.5VL-32B~\cite{qwen25vl}          & 0.45 & 6.42 & -0.50 & 0.67 & -0.03 & 0.73 & 81.73 & 84.09 & 80.06 & \textcolor{red}{\ding{55}} \\
 & Qwen2.5VL-72B~\cite{qwen25vl}          & 0.87 & 4.22 & 0.50 & 0.67 & 0.10 & 0.34 & 75.36 & 81.69 & 81.66 & \textcolor{red}{\ding{55}} \\
 & LLaVA-OV-72B~\cite{li2024llavaonevision}             & 0.03 & 9.71 & 0.50 & 0.67 & 0.02 & 0.40 & 79.59 & 79.46 & 80.63 & \textcolor{red}{\ding{55}} \\
 & Gemma-3-27B~\cite{gemma3}              & 1.01 & 3.70 & 0.50 & 0.67 & 0.09 & 0.41 & 75.39 & 74.55 & 80.83 & \textcolor{red}{\ding{55}} \\
\midrule

\multirow{4}{*}{\textbf{Closed Source}}
 & Gemini-2.5 Pro~\cite{gemini25pro}      & 3.31 & 0.42 & 1.00 & 0.00 & 0.25 & 0.18 & 52.90 & 64.00 & 67.70 & \textcolor{green}{$\checkmark$} \\
 & Claude Sonnet 4~\cite{claude4sonnet}    & 2.09 & 1.32 & 1.00 & 0.00 & 0.16 & 0.08 & 67.36 & 71.12 & 76.94 & \textcolor{green}{$\checkmark$} \\
 & GPT-4o~\cite{gpt4o}                    & 1.63 & 2.04 & 1.00 & 0.00 & 0.16 & 0.13 & 76.92 & 83.48 & 86.63 & \textcolor{green}{$\checkmark$} \\
 & o3~\cite{o3}                            & 9.84 & 0.02 & 1.00 & 0.00 & 0.40 & 0.10 & 59.27 & 74.47 & 83.18 & \textcolor{green}{$\checkmark$} \\
\bottomrule
\end{tabular}
}%
\label{tab:anova-trend}
\end{table*}

As shown in Figure~\ref{fig:PointDis} and Table~\ref{tab:data-distribution}, the global distribution of WanderBench locations also induces rich variation in spatial diversity metrics computed over model proposed coordinates. Models that spread their predictions broadly across continents exhibit higher occupancy, entropy, and convex-hull area, indicating that they explore many distinct subregions instead of collapsing onto a few high density areas. For example, o3 attains the largest hull area ($A_{\text{hull}}{=}12.8{\times}10^{-2}$) and one of the highest Clark Evans ratios ($R_{\text{CE}}{=}3.70{\times}10^{-1}$), signaling wide geographical coverage and relatively uniform dispersion, while InternVL2-76B and Gemini~2.5~Pro show similarly strong entropy and occupancy values. In contrast, weaker or more region-biased models such as Qwen2VL-72B and Claude Sonnet~4 exhibit smaller hull areas and $R_{\text{CE}}{<}1$, revealing tightly clustered predictions concentrated in familiar regions. These spatial diversity statistics, together with the underlying graph structure, demonstrate that WanderBench is well suited for assessing both geolocation accuracy and global coverage behavior. Moreover, because these metrics expose subtle regional biases and mode collapse tendencies, they provide a complementary lens for interpreting failure cases and distinguishing models that merely localize well from those that truly generalize across the full geographic spectrum.

\section{Experiment}
\subsection{Evaluation Setup}

We conduct comprehensive experiments on 20 leading LMMs with various architectures and parameter scales on our WanderBench. Those models are categorized into two groups: (1) open sourced models, including InternVL series \cite{internvl2}, Qwen-VL series \cite{qwen25vl},
LLaVA-OneVision \cite{li2024llavaonevision}, \textit{etc}. (2)close sourced models, including GPT-4o \cite{gpt4o}, o3 \cite{o3}, Gemini \cite{comanici2025gemini}, \textit{etc}. The proprietary models are all directly acquired from offical API providers and all opensource model are deploy on local with ms swift framework on H200 GPUs. All models are evaluated using the same prompt template provided in the Appendix. To ensure reproducibility, all models are set the temperature to 0 and perform greedy decoding.

Noticeably, model performance on the Geolocation will also be reflected by its ability to propose valid and diverse locations and performance with different difficulties. Therefore, unlike the pervious benchmarks, WounderBench further evaluate the location proposing capability of those models. For this purpose, we introduce additional evaluation metrics specifically designed to assess the quality and diversity of the proposed locations. The evaluation metrics and results are provided in the following sections.

\begin{figure*}
    \includegraphics[width=\textwidth]{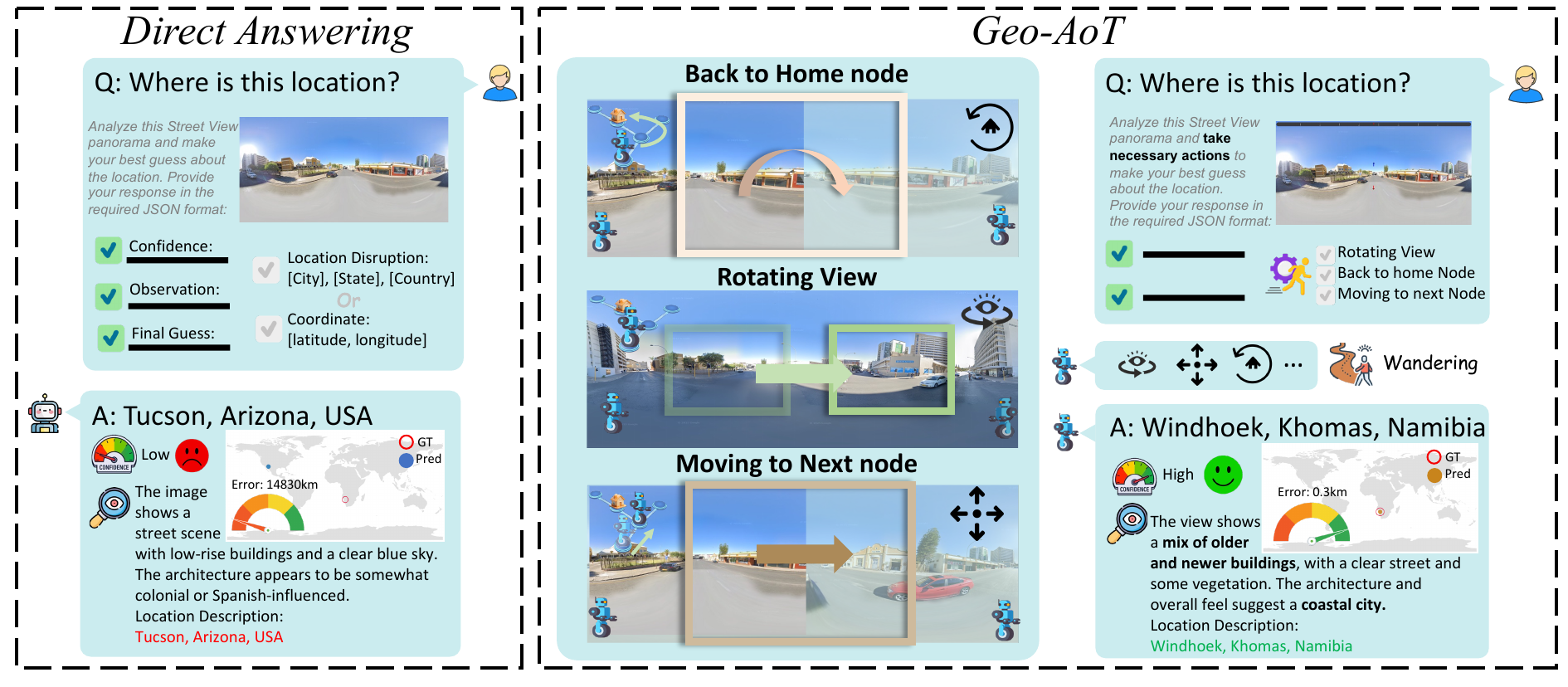}
\caption{An overview of GeoAoT. Given an input image, GeoAoT first leverages a pre-trained LMMs to generate an initial geo-guess. It then iteratively refines this estimate through AoT based multi-turn interactions, where the model reasons about uncertainty and issues actions to gather more evidence, operating purely at inference time without any additional training.}
\label{fig:Geo-AoT}
\end{figure*}

\subsection{Geo-Guessing Capability Evaluation}

Table~\ref{tab:baseline_results} presents a comprehensive comparison of our model with both open- and closed-source Large Multimodal Models (LMMs) on the geo-guessing benchmark. We assess performance across three hierarchical spatial levels:\textit{street}, \textit{city}, and \textit{country} and report standard classification metrics (accuracy, recall, F1) alongside great circle distance error and Geo Score:
\begin{align}
    Score = 100 \cdot \exp\!\left(-\frac{10\,x}{18050}\right),
\end{align}
which align with \cite{geoguessr}, providing a holistic measurement of fine- and coarse-grained geolocation ability. This multi-level evaluation captures not only a model's capability to recognize broad geographic descriptors such as climate and terrain but also its sensitivity to localized cues such as architectural motifs, signage, and road patterns.

The results reveal clear trends in model behavior. Open source LMMs, particularly the Qwen2.5-VL and Qwen2-VL families, demonstrate rapid progress and now compete closely with closed source frontier systems, achieving strong city- and country-level performance while maintaining reasonable street level accuracy. Retrieval based models like GeoCLIP exhibit strength in coarse grained localization due to their exposure to large-scale image–GPS pretraining but struggle to distinguish finer geographic details, resulting in diminished street- and city-level accuracies. Meanwhile, larger open source LMMs including InternVL variants, LLaMA based models, and GLM-4.5V show varying trade-offs between granularity and overall stability, with GLM-4.5V standing out as a particularly strong open-source performer across all metrics.

Closed-source frontier models achieve the best overall results. Gemini-2.5-Pro attains the lowest distance error (134.32 km) and highest composite score (95.28), outperforming all models at both fine- and coarse-level prediction. Gemini-2.5-Flash, GPT-4o, and o3 also deliver consistently strong accuracy and recall across all spatial resolutions, indicating robust spatial reasoning and effective integration of visual and world knowledge. Taken together, these findings highlight continued advancement in geolocation capabilities across the LMM landscape and underscore the narrowing gap between leading open- and closed-source systems in global image geolocation.

\subsubsection{ANOVA and Trend Analysis}

To assess robustness across coverage levels (20\%, 50\%, 80\%), we conduct a one-way ANOVA and monotonic trend analysis as shown in Table \ref{tab:anova-trend}. The updated ANOVA results show that only a subset of models exhibits significant variance across conditions. Notably, InternVL2-76B ($F=7.62$, $p=0.001$), Gemini-2.5 Pro ($F=3.31$, $p=0.004$), and o3 ($F=9.84$, $p=0.0002$) display clear coverage-dependent performance shifts, while many others including several Qwen2/2.5 and InternVL2.5 models show nonsignificant differences, indicating flatter or less stable responses to increased coverage.
\begin{table*}[t]
\caption{Performance of GeoAoT on the benchmark. \textcolor{blue}{\textbf{Blue bold}} values represent the best AoT result. \textcolor{red}{\textbf{Red bold}} values represent the strongest improvement per column.}
\centering
\setlength{\tabcolsep}{3pt}
\renewcommand{\arraystretch}{0.95}
\resizebox{\textwidth}{!}{%
\begin{tabular}{
    >{\raggedright\arraybackslash}m{4.6cm}
    *{12}{>{\centering\arraybackslash}m{1.1cm}}
}
\toprule
\multirow{2}{*}{\shortstack[c]{\textbf{Method}}} &
\multicolumn{2}{c}{\textbf{Overall}} &
\multicolumn{3}{c}{\textbf{Street}} &
\multicolumn{3}{c}{\textbf{City}} &
\multicolumn{3}{c}{\textbf{Country}} \\
\cmidrule(lr){2-3}\cmidrule(lr){4-6}\cmidrule(lr){7-9}\cmidrule(lr){10-12}
& \textbf{Dist.\,$\downarrow$} & \textbf{Score\,$\uparrow$} &
\textbf{Acc.\,$\uparrow$} & \textbf{Rec.\,$\uparrow$} & \textbf{F1\,$\uparrow$} &
\textbf{Acc.\,$\uparrow$} & \textbf{Rec.\,$\uparrow$} & \textbf{F1\,$\uparrow$} &
\textbf{Acc.\,$\uparrow$} & \textbf{Rec.\,$\uparrow$} & \textbf{F1\,$\uparrow$} \\
\midrule
LLaVA OneVision 72B
& 4081 & 53.68 & 0.112 & 0.097 & 0.104 & 0.241 & 0.206 & 0.222 & 0.508 & 0.482 & 0.494 \\
(w/ AoT)
& 3007 & 61.20 & 0.150 & 0.130 & 0.139 & 0.304 & 0.261 & 0.281 & 0.579 & 0.544 & 0.561 \\
$\Delta$ (AoT $-$ base)
& $\textcolor{red}{\textbf{-1074}}$ & $\textcolor{red}{\textbf{7.527}}$ & $0.038$ & $0.033$ & $0.035$ & $0.063$ & $0.055$ & $0.058$ & $\textcolor{red}{\textbf{0.071}}$ & $\textcolor{red}{\textbf{0.062}}$ & $\textcolor{red}{\textbf{0.067}}$ \\
\midrule
Gemma3 27B
& 1685 & 72.76 & 0.207 & 0.178 & 0.191 & 0.426 & 0.366 & 0.393 & 0.685 & 0.681 & 0.683 \\
(w/ AoT)
& 1475 & 74.95 & 0.228 & 0.196 & 0.211 & 0.442 & 0.380 & 0.409 & 0.711 & 0.704 & 0.708 \\
$\Delta$ (AoT $-$ base)
& $-209.9$ & $2.191$ & $0.021$ & $0.018$ & $0.019$ & $0.016$ & $0.014$ & $0.015$ & $0.027$ & $0.023$ & $0.025$ \\
\midrule
Qwen2.5-VL 72B
& 674.8 & 84.58 & 0.257 & 0.221 & 0.238 & 0.55 & 0.473 & 0.509 & 0.833 & 0.810 & 0.821 \\
(w/ AoT)
& 522.4 & 87.02 & 0.270 & 0.233 & 0.250 & 0.584 & 0.502 & 0.540 & 0.863 & 0.835 & 0.849 \\
$\Delta$ (AoT $-$ base)
& $-152.3$ & $2.441$ & $0.013$ & $0.011$ & $0.012$ & $0.033$ & $0.029$ & $0.031$ & $0.030$ & $0.025$ & $0.027$ \\
\midrule
InternVL3 78B
& 1442 & 72.77 & 0.171 & 0.148 & 0.159 & 0.359 & 0.309 & 0.332 & 0.697 & 0.646 & 0.670 \\
(w/ AoT)
& 1253 & 73.95 & 0.177 & 0.153 & 0.164 & 0.371 & 0.319 & 0.343 & 0.710 & 0.658 & 0.683 \\
$\Delta$ (AoT $-$ base)
& $-189.1$ & $1.181$ & $0.006$ & $0.005$ & $0.005$ & $0.012$ & $0.011$ & $0.012$ & $0.013$ & $0.012$ & $0.013$ \\
\midrule
o3
& 318.8 & 92.01 & 0.524 & 0.525 & 0.525 & 0.734 & 0.734 & 0.734 & 0.922 & 0.921 & 0.922 \\
(w/ AoT)
& \textcolor{blue}{\textbf{180.3}} & \textcolor{blue}{\textbf{94.85}} &
  \textcolor{blue}{\textbf{0.606}} & \textcolor{blue}{\textbf{0.608}} & \textcolor{blue}{\textbf{0.607}} &
  \textcolor{blue}{\textbf{0.810}} & \textcolor{blue}{\textbf{0.810}} & \textcolor{blue}{\textbf{0.810}} &
  \textcolor{blue}{\textbf{0.950}} & \textcolor{blue}{\textbf{0.950}} & \textcolor{blue}{\textbf{0.950}} \\

$\Delta$ (AoT $-$ base)
& $-138.5$ & $2.841$ &
  $\textcolor{red}{\textbf{0.082}}$ & $\textcolor{red}{\textbf{0.082}}$ & $\textcolor{red}{\textbf{0.082}}$ &
  $\textcolor{red}{\textbf{0.076}}$ & $\textcolor{red}{\textbf{0.076}}$ & $\textcolor{red}{\textbf{0.076}}$ &
  $0.029$ & $0.029$ & $0.029$ \\
\midrule
Gemini 2.5 flash
& 222.1 & 93.37 & 0.504 & 0.505 & 0.505 & 0.742 & 0.742 & 0.742 & 0.941 & 0.940 & 0.941 \\
(w/ AoT)
& \textcolor{red}{\textbf{166.2}} & 94.18 & 0.510 & 0.511 & 0.510 & 0.753 & 0.753 & 0.753 & 0.949 & 0.949 & 0.949 \\
$\Delta$ (AoT $-$ base)
& $-55.83$ & $0.818$ & $0.006$ & $0.005$ & $0.006$ & $0.011$ & $0.010$ & $0.010$ & $0.009$ & $0.009$ & $0.009$ \\
\midrule
Claude Haiku 4.5
& 2715 & 59.54 & 0.134 & 0.134 & 0.134 & 0.268 & 0.268 & 0.268 & 0.538 & 0.537 & 0.537 \\
(w/ AoT)
& 2520 & 62.01 & 0.145 & 0.145 & 0.145 & 0.279 & 0.279 & 0.279 & 0.563 & 0.562 & 0.562 \\
$\Delta$ (AoT $-$ base)
& $-195.0$ & $2.468$ & $0.011$ & $0.012$ & $0.012$ & $0.011$ & $0.010$ & $0.010$ & $0.025$ & $0.025$ & $0.025$ \\
\bottomrule
\end{tabular}%
}
\label{tab:AoT_results}
\end{table*}

For monotonicity, we examine Spearman correlations and OLS slopes. A majority of higher-performing models, including InternVL2-76B, InternVL2.5-38B, InternVL3-38B/78B, Phi-4, Claude Sonnet 4, Gemini-2.5 Pro, GPT-4o, and o3, achieve perfect monotonic trends ($\rho = 1.0$), consistently improving as coverage increases. In contrast, the Qwen2 and Qwen2.5 variants exhibit weak or inconsistent monotonicity, often showing performance dips at higher coverage levels.

Overall, ten models satisfy both statistical and monotonic consistency among them GPT-4o, o3, Gemini-2.5 Pro, and multiple InternVL models indicating stronger generalization as data availability grows. Meanwhile, several large open source models display non-monotonic or noisy trajectories, suggesting potential instability or limited robustness under varying coverage.
\section{GeoAoT}

\subsection{Framework Overview}

GeoAoT reformulates the traditional \textit{Chain-of-Thought} (CoT) paradigm into an \textit{Action-of-Thought} (AoT) process. As illustrated in Figure~\ref{fig:Geo-AoT}, the agent operates in an iterative loop composed of two stages: \textit{geo guessing} and \textit{action taking}. In each iteration, the LMM first produces an initial geolocation estimate based on the current visual observation. It then reflects on the uncertainty of its reasoning identifying which visual cues are missing or ambiguous and formulates a concrete action to gather additional evidence.

Formally, the GeoAoT process is expressed as:
\begin{equation}
o_t = f_{\text{LMM}}(I_t, h_t);
a_t = \pi_{\text{AoT}}(o_t);
I_{t+1} = \mathcal{T}(I_t, a_t),  \label{eq:perception_action_loop}
\end{equation}
where $I_t$ denotes the current observation, $f_{\text{LMM}}$ the model’s multimodal reasoning step, $\pi_{\text{AoT}}$ the action policy derived from textual reasoning, and $\mathcal{T}$ the environment transition function that executes the proposed action.

Unlike static CoT reasoning, GeoAoT explicitly grounds each reasoning step in a corresponding environmental action. For example, if the model determines that “the signage is unreadable,” the AoT framework converts this into the actionable instruction “rotate 30° right.” This coupling allows the model to interact with the environment, reducing uncertainty through iterative feedback rather than relying solely on single-image inference.

The AoT framework uses structured prompting templates that guide the LMM to (i) analyze visible geographical and cultural cues, (ii) identify missing or ambiguous evidence, (iii) propose an environmental action to reduce uncertainty, and (iv) update its geolocation hypothesis based on the new observation.

\subsection{Evaluation and Observations}

GeoAoT consistently improves geolocation performance across all evaluated LMMs (Table~\ref{tab:AoT_results}). Every model benefits from reduced distance error and higher accuracy metrics, with overall error decreasing from 56,km up to more than 1{,}000,km depending on model capacity. High performing systems such as o3 and Gemini2.5Flash achieve notable error reductions (138.5 km and 55.8 km) alongside the largest fine grained gains, while weaker baselines like LLaVA OneVision 72B experience the strongest absolute improvements ($-1074$ km). These results show that iterative observation benefits both robust and weaker models, systematically enhancing localization reliability.

Improvements appear consistently across spatial scales. Street level F1 scores rise for all models, from +0.005 (InternVL3~78B) to +0.082 (o3), with similarly stable gains at the city level (+0.010 to +0.076). Even at the already strong country level, GeoAoT yields measurable boosts (e.g., +0.029 for o3, +0.025 for Claude Haiku). Qualitatively, the method produces more interpretable reasoning traces: instead of issuing a single opaque prediction, models perform stepwise analysis grounded in explicit observational actions. This interaction supported reasoning reduces ambiguity and offers clearer insight into how predictions evolve, demonstrating that GeoAoT improves both accuracy and transparency.
\section{Conclusion}
We introduced \textbf{WanderBench}, the first open-access global geolocation benchmark enabling embodied actions and bidirectional evaluation, and \textbf{GeoAoT}, a framework that couples multimodal reasoning with active exploration. Together, they reframe geolocation from static prediction to interactive inference across 32K panoramas spanning six continents, providing a unified testbed for both localization accuracy and model driven task generation. Experiments on 19 large multimodal models show that AoT based exploration consistently improves fine-grained localization and produces more interpretable reasoning aligned with physical actions. Beyond performance gains, our bidirectional protocol evaluates a model’s ability not only to solve geolocation tasks but also to construct diverse, difficulty calibrated ones, enabling deeper assessment of spatial understanding. These contributions establish a new foundation for embodied geolocation research and point toward future LMMs capable of more autonomous, self-evaluative, and geographically grounded intelligence.
\bibliographystyle{ieeenat_fullname}
\bibliography{references}

\end{document}